\documentclass[10pt,twocolumn,letterpaper]{article}

\usepackage{cvpr}
\usepackage{times}
\usepackage{epsfig}
\usepackage{graphicx}
\usepackage{amsmath}
\usepackage{amssymb}

\usepackage{color}
\usepackage{comment}
\usepackage{etoolbox}
\usepackage{booktabs}
\usepackage{multirow}

\usepackage{caption}
\usepackage{subcaption}

\usepackage{bm}
\usepackage[pagebackref=true,breaklinks=true,letterpaper=true,colorlinks,bookmarks=false]{hyperref}

\usepackage{xcolor}
\newbool{inccomment}
\setbool{inccomment}{true}
\newcommand{\pk}[1]{\ifbool{inccomment}{{\color{magenta}#1}}{}}
\newcommand{\ww}[1]{\ifbool{inccomment}{{\color{red} #1}}{}}
\newcommand{\gb}[1]{\ifbool{inccomment}{{\color{blue} #1}}{}}
\newcommand{\hl}[1]{\ifbool{inccomment}{{\color{orange} #1}}{}}

 \cvprfinalcopy % *** Uncomment this line for the final submission

\def\cvprPaperID{9829} % *** Enter the CVPR Paper ID here

\ifcvprfinal\fi
\begin{document}

%%%%%%%%% TITLE
\title{Neural Predictor for Neural Architecture Search}

\author{Wei Wen\\
Google Brain and Duke University\\
{\tt\small wei.wen@duke.edu}
\and
Hanxiao Liu\\
Google Brain\\
{\tt\small hanxiaol@google.com}
\and
Hai Li\\
Duke University\\
{\tt\small hai.li@duke.edu}
\and
Yiran Chen\\
Duke University\\
{\tt\small yiran.chen@duke.edu}
\and
Gabriel Bender\\
Google Brain\\
{\tt\small gbender@google.com}
\and
Pieter-Jan Kindermans\\
Google Brain\\
{\tt\small pikinder@google.com}
}
\maketitle
%\thispagestyle{empty}
%%%%%%%%% ABSTRACT
\begin{abstract}
Neural Architecture Search methods are effective but often use complex algorithms to come up with the best architecture. We propose an approach with three basic steps that is conceptually much simpler. First we train $N$ random architectures to generate $N$ (architecture, validation accuracy) pairs and use them to train a regression model that predicts accuracy based on the architecture. Next, we
use this regression model to predict the validation accuracies of a large number of random architectures. Finally, we train the top-$K$ predicted architectures and deploy the model with the best validation result. While this approach seems simple, it is more than $20 \times$ as sample efficient as Regularized Evolution on the NASBench-101 benchmark and can compete on ImageNet with more complex approaches based on weight sharing, such as ProxylessNAS.
\end{abstract}

%RESOLVED \gb{Can we please use a word other than \textbf{simple} to describe our method throughout the paper? It's debatable whether graph networks are simple. (It's debatable, but personally, I don't feel comfortable describing them as simple.) It's still significant that we're able to improve upon techniques like RL, evolution, and active learning using a classical supervised ML setup plus random sampling, and I think that's the part to emphasize.}

%%%%%%%%% BODY TEXT
\section{Introduction}
The original Neural Architecture Search (NAS) methods have resulted in improved accuracy but they came at a high computational cost \cite{zoph2018learning,real2017large,real2019regularized}. Recent advances have reduced this cost significantly \cite{liu2018darts,hu2019efficient,zhou2019bayesnas,cai2018proxylessnas,pham2018efficient,dai2019chamnet,bender2018understanding,luo2018neural,xie2018snas,wu2019fbnet,brock2017smash},
but
%it remains unclear how many of these methods could be applied directly to a complex search space such as the one used by NASBench-101 \cite{ying2019bench} as
many of them require nontrivial specialized implementations. For example, weight sharing introduces additional complexity into the search process, and must be carefully tuned to get good results.

%\hl{Is it possible to find some other ways to motivate this? My impression is that search spaces in some of these works are at least equally challenging as NASBench-101. Shall we just say ``but many of these methods require nontrivial specialized implementations and/or costly hyperparameter tuning processes''?}.
% these methods might still require strict hyper-parameter tuning. Also, their effectiveness over random search has been questioned by several papers (such as Li~\textit{et al.}~\cite{li2019random}) and they can be conceptually quite complex.

With an infinite compute budget, a na\"ive approach to architecture search would be to sample tens or hundreds of thousands of random architectures, train and evaluate each one, and then select the architectures with the best validation set accuracies for deployment; this is a straightforward application of the ubiquitous \emph{random search} heuristic. However, the computational requirements of this approach makes it infeasible in practice. For example, to exhaustively train and evaluate each of the 400,000 architectures in the NASBench~\cite{ying2019bench} search space, it would take roughly 25 years of TPU training time.
Only a small number of companies and corporate research labs can afford this much compute, and it is far out of reach for most ML practitioners.

% With an infinite compute budget, exhaustive search would be able to do exactly what every machine learning student is taught. To select the optimal configuration, train all possible models in the search space (i.e. hyper-parameters or architectures) and deploy the model with the best validation accuracy. For nontrivial real-world search spaces, the computational requirements make this approach infeasible. For CIFAR-10, this would take 25 years of TPU training time if we limit ourselves to architectures in the NASBench-101 dataset~\cite{ying2019bench} -- a search space which was specifically engineered to be exhaustively searchable. It goes without saying that only the largest companies can afford so much compute and it is far out of reach for regular ML practitioners.

\begin{figure}
\begin{center}
\includegraphics[width=\columnwidth]{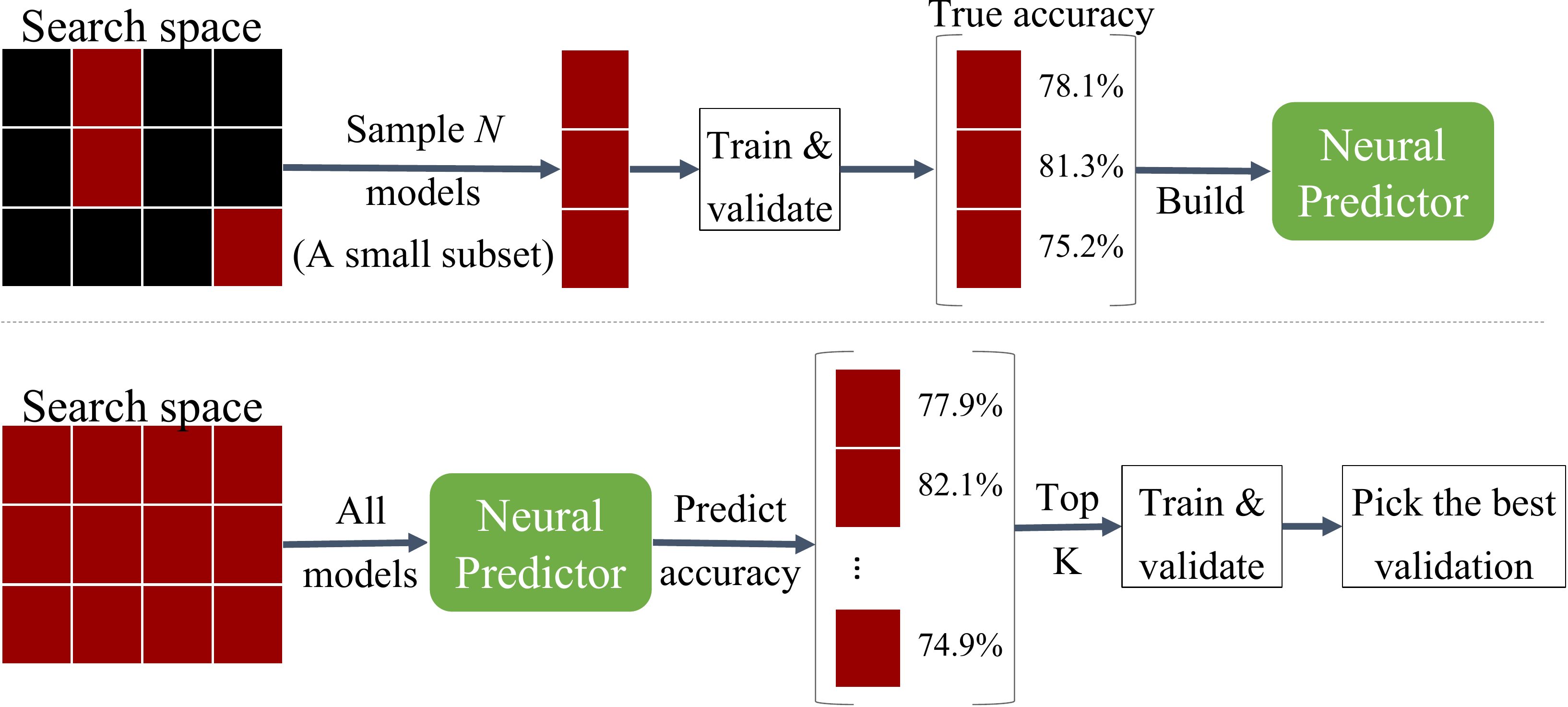}
\end{center}
  \caption{Training and applying the Neural Predictor. %\hl{The figure looks nice but the texts are too small to read. Is it possible to enlarge them?}}
  }
\label{fig:build_use_np}
\end{figure}

One way to alleviate this is to identify a small subset of promising models. If we can do this with a reasonably high recall (most models selected are indeed of high quality) then we can train and validate just this limited set of models to reliably select a good one for deployment. To achieve this, the proposed Neural Predictor uses the following steps to perform an architecture search:

\textbf{(1) Build a predictor} by training $N$ random architectures to obtain $N$ (architecture, validation accuracy) pairs. Use this data to train a regression model.

\textbf{(2) Quality prediction} using the regression model over a large set of random architectures. Select the $K$ most promising architectures for final validation.

\textbf{(3) Final validation} of the top $K$ architectures by training them. Then we select the model with the highest validation accuracy to deploy. 

The workflow is illustrated in Figure~\ref{fig:build_use_np}.
In this setup, the first step is a traditional regression problem where we first generate a dataset of $N$ samples to train on. The second step can be carried out efficiently because evaluating a model using the predictor is cheap. The third step is nothing more than traditional validation where we only evaluate a well curated set of $K$ models.
% Here, training the $N$ initial models and the final validation on $K$ models are the expensive steps.
While the method outlined above might seem straightforward, it is very effective:
\begin{itemize}
\item The Neural Predictor strongly outperforms random search on NASBench-101. It is also about $22.83$ times more sample-efficient than Regularized Evolution, the best performing method in the NASBench-101 paper.
\item The Neural Predictor can easily handle different search spaces. In addition to NASBench-101, we evaluated it on the ProxylessNAS \cite{cai2018proxylessnas} search space and found that the predicted architecture is as accurate as ProxylessNAS and clearly better than random search.
\item The architecure selection process uses two of the most ubiquitous tools from the ML toolbox: random sampling and supervised learning. In contrast, many existing NAS approaches rely on reinforcement learning, weight sharing, or Bayesian optimization.
% The major complexity lies in the setup of the regression model. Despite this, it is conceptually much simpler than typical NAS approaches that rely on reinforcement learning, weight sharing or Bayesian optimization.

\item The most computationally intensive components of the proposed method (training $N$ models in step 1 and $K$ models in step 3) are highly parallelizable when sufficient computation resources are available.

% \item If sufficient compute resources are available, most steps of the Neural Predictor can be parallelized well.
\end{itemize}

\section{Neural Predictor}

\begin{figure*}
\begin{center}
\includegraphics[width=.95\textwidth]{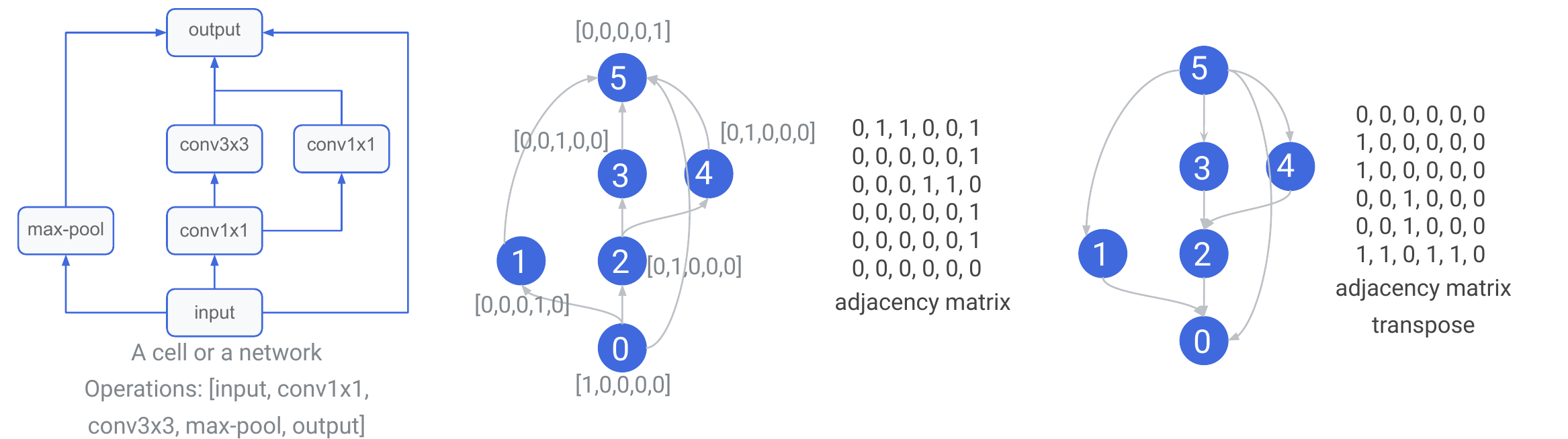}
\vspace{-12pt}
\end{center}
  \caption{An illustration of graph and node representations. Left: A neural network architecture with $5$ candidate options per node, each represented as a one-hot code. The one-hot codes are inputs of a bidirectional GCN, which takes into account both the original adjacency matrix (middle) and its transpose (right).}
\label{fig:GCN_example}
\end{figure*}

The core idea behind the Neural Predictor is that carrying out the actual training and validation process is the most reliable way to find the best model. The goal of the Neural Predictor is to provide us with a curated list of promising models for final validation prior to deployment. The entire Neural Predictor process is outlined below. 

\textbf{Step 1: Build the predictor using $N$ samples.}
We train $N$ models to obtain a small dataset of (architecture, validation accuracy) pairs.
The dataset is then used to train a regression model that maps an architecture to the predicted validation accuracy.

\textbf{Step 2: Quality prediction.}
Because architecture evaluation using the learned predictor is efficient and trivially parallelizable, we use it to rapidly predict the accuracies of a large number of random architectures. We then select the top $K$ predicted architectures for final validation.

\textbf{Step 3: Final validation on $K$ samples.}
We train and validate the top $K$ models in the traditional way. This allows us to select the best model based on the actual validation accuracy. Even if our predictor is somewhat noisy, this step allows us to use a more reliable measurement to select our final architecture for deployment.  % A key advantage of this step is that even if our predictor is somewhat noisy, our final selection for deployment is based on a more reliable measurement.

Training $N$+$K$ models is by far the most computationally expensive part of the Neural Predictor. If we assume a constant compute budget, $N$ and $K$ are key hyper-parameters which needs to be set; we will discuss this next. 

\subsection{Hyper-parameters in the Workflow}
\textbf{Hyper-parameters for (final) model training} are always needed if we train a single model or we perform an architecture search. In this respect the Neural Predictor is no different from other methods. We found that using the same hyper-parameters for all models we train is an effective strategy, one that was also used in NASBench-101. 

\textbf{Trade-off between $N$ and $K$ for a fixed budget:}
If we increase $N$, the number of samples used to train the Neural Predictor, we can expect the predictor to become more accurate. However, to maintain a fixed compute budget, we must decrease $K$ in order to increase $N$.

If $K$ is large and the predictor proposes a good set of architectures for final validation, the precise ranking of these architecture is not that important. 

However, if we use more training data to improve the predictor, we have a small set of $K$ models for final evaluation. In this case, the performance predictor must be able to reliably identify high-quality models from the search space. This might be a hard task and it is possible that trying out more models is more effective than using more data to get a small improvement in predictive quality. Because it is difficult to theoretically predict the optimal trade-off, we will investigate this in the experimental setting. 

To find a lower bound on $N$ we could iteratively increase $N$ until we observe a good cross-validation accuracy. This means that contrast to some other methods such as Regularized Evolution~\cite{real2019regularized}, ENAS~\cite{pham2018efficient}, NASNet~\cite{zoph2018learning}, ProxylessNAS~\cite{cai2018proxylessnas}, there is no need to repeat the entire search experiment in order to tune this hyper-parameter. The same applies to the hyper-parameters and the architecture of the Neural Predictor itself.

\textbf{The hyper-parameters of the Neural Predictor} can be optimized by cross-validation using the $N$ training samples. In contrast, RL or Evolution-based search methods would require us to collect more training samples in order to try out a new hyper-parameter configuration. We tried many options for the architecture of the predictor. Due to space constraints we will limit our discussion to Graph Convolutional Networks. A comparison against other regression methods can be found in the supplementary material.

\subsection{Modeling by Graph Convolutional Networks}
Graph Convolutional Networks (GCNs) are good at learning representations for graph-structured data~\cite{kipf2016semi, velivckovic2017graph} such as a neural network architecture. 
The graph convolutional model we use is based on \cite{kipf2016semi}, which assumes undirected graphs. We will modify their approach to handle neural architectures represented as directed graphs.

%\gb{Is A meant to be an adjacency matrix? If so, I think it would be helpful to say so explicitly in the paragraph below.}
We start with a $D_0$-dimensional representation for each of the $I$ nodes in the graph, giving us an initial feature vector $\bm{V}_0 \in \mathbb{R}^{I \times D_0}$. For each node we use a one-hot vector representing the selected operation. An example for NASBench-101 is shown in Figure~\ref{fig:GCN_example}. 
The node representation is iteratively updated using Graph Convolutional Layers. Each layer uses an adjacency matrix $\bm{A} \in \mathbb{R}^{I \times I}$ based on the node connectivity and a trainable weight matrix $\bm{W}_l \in \mathbb{R} ^ {D_l \times D_{l+1}}$:
\begin{equation}
\label{eq:gc}
    \bm{V}_{l+1} = \mathrm{ReLU} \left( \bm{A} \bm{V}_l \bm{W}_l \right).
\end{equation}
Following \cite{kipf2016semi}, we add an identity matrix to $\bm{A}$ (corresponding to self cycles) and normalize it using the node degree.

The original GCNs \cite{kipf2016semi} assume undirected graphs.
% When applied to a directed acyclic graph the directed adjacency matrix the information flows only in a single direction.
When applied to a directed acyclic graph, the directed adjacency matrix allows information to flow only in a single direction.
To make information flow both ways, we always use the average of two GCN layers: one where we use $\mathbf{A}$ to propagate information in the forward directions and another where we use $\mathbf{A}^T$ to reverse the direction:
\begin{equation*}
    \bm{V}_{l+1} = \frac{1}{2}\mathrm{ReLU} \left( \bm{A} \bm{V}_l \bm{W}^+_l \right)+\frac{1}{2}\mathrm{ReLU} \left( \bm{A}^T \bm{V}_l \bm{W}^-_l \right).
\end{equation*}
Figure~\ref{fig:GCN_example} shows an example of how the adjacency matrices are constructed (without normalization or self-cycles).

GCNs are able to learn high quality node representations by stacking multiple of these layers together. Since we are more interested in the accuracy of the overall network (a global property), we take the average over node representations from the final graph convolutional layer and attach one or more fully connected layers to obtain the desired output. Details are provided in the supplementary material.
\section{Experiments}
In this section we will discuss two studies. First we will analyze the Neural Predictor's behavior in the controlled environment from NASBench-101~\cite{ying2019bench}. Afterwards we will use our approach to search for high quality mobile models in the ProxylessNAS search space~\cite{cai2018proxylessnas}.

% \begin{enumerate}
%     \item 90 - 360 for all frontiers
%     \item simplicity
%     \item N-K trade-off
% %    \item colors http://colorbrewer2.org/#type=qualitative&scheme=Paired&n=10
%     \item log scale of number of samples
%     \item a single run time for NP and proxylessNAS
% \end{enumerate}

\subsection{NASBench-101}
NASBench-101 \cite{ying2019bench} is a dataset used to benchmark NAS algorithms. The goal is to come up with a high quality architecture as efficiently as possible.
The dataset has the following properties.
\textbf{(1)} Train time, validation and test results are provided for all 423,624 models in the search space. 
\textbf{(2)} Each model was trained and evaluated three times. This allows us to look at the variance across runs.% caused by a potentially different initialization.
\textbf{(3)} All models were trained in a consistent manner, preventing biases from the implementation from skewing results.
\textbf{(4)} NASBench-101 recommends using only validation accuracies during a search, and reserving test accuracies in the final report; this is important to avoid overfitting.

NASBench-101 uses a cell-based NAS~\cite{zoph2018learning} on CIFAR-10~\cite{krizhevsky2009learning}. Each cell is a Directed Acyclic Graph (DAG) with an input node, an output node and up to $5$ interior nodes. Each interior node can be a $1 \times 1$ convolution (\texttt{conv1x1}), $3 \times 3$ convolution (\texttt{conv3x3}) or max-pooling op (\texttt{max-pool}). One example is shown in Figure~\ref{fig:GCN_example} (left).

In each experiment, we use the validation accuracy from a \textit{single run}\footnote{In the training dataset of our Neural Predictor, this means that each model's accuracy label is sampled once and fixed across all epochs.} as a search signal. The single run is uniformly sampled from these three records. This simulates training the architecture once. Test accuracy is only used for reporting the accuracy on the model that was selected at the end of a search. For that model we use the \textit{mean} test accuracy over three runs as the ``ground truth'' measure of accuracy.

% \begin{figure}
% \centering
% \begin{minipage}{.5\columnwidth}
%   \centering
%   \includegraphics[width=1.\columnwidth]{figures/oracle_val_vs_test.png}
%   \captionof{figure}{A figure}
%   \label{fig:test1}
% \end{minipage}%
% \begin{minipage}{.5\columnwidth}
%   \centering
%   \includegraphics[width=1.\columnwidth]{figures/oracle_val_vs_test_zoom_in.png}
%   \captionof{figure}{Another figure}
%   \label{fig:test2}
% \end{minipage}
% \end{figure}

\begin{figure}
\begin{center}
\includegraphics[width=\columnwidth]{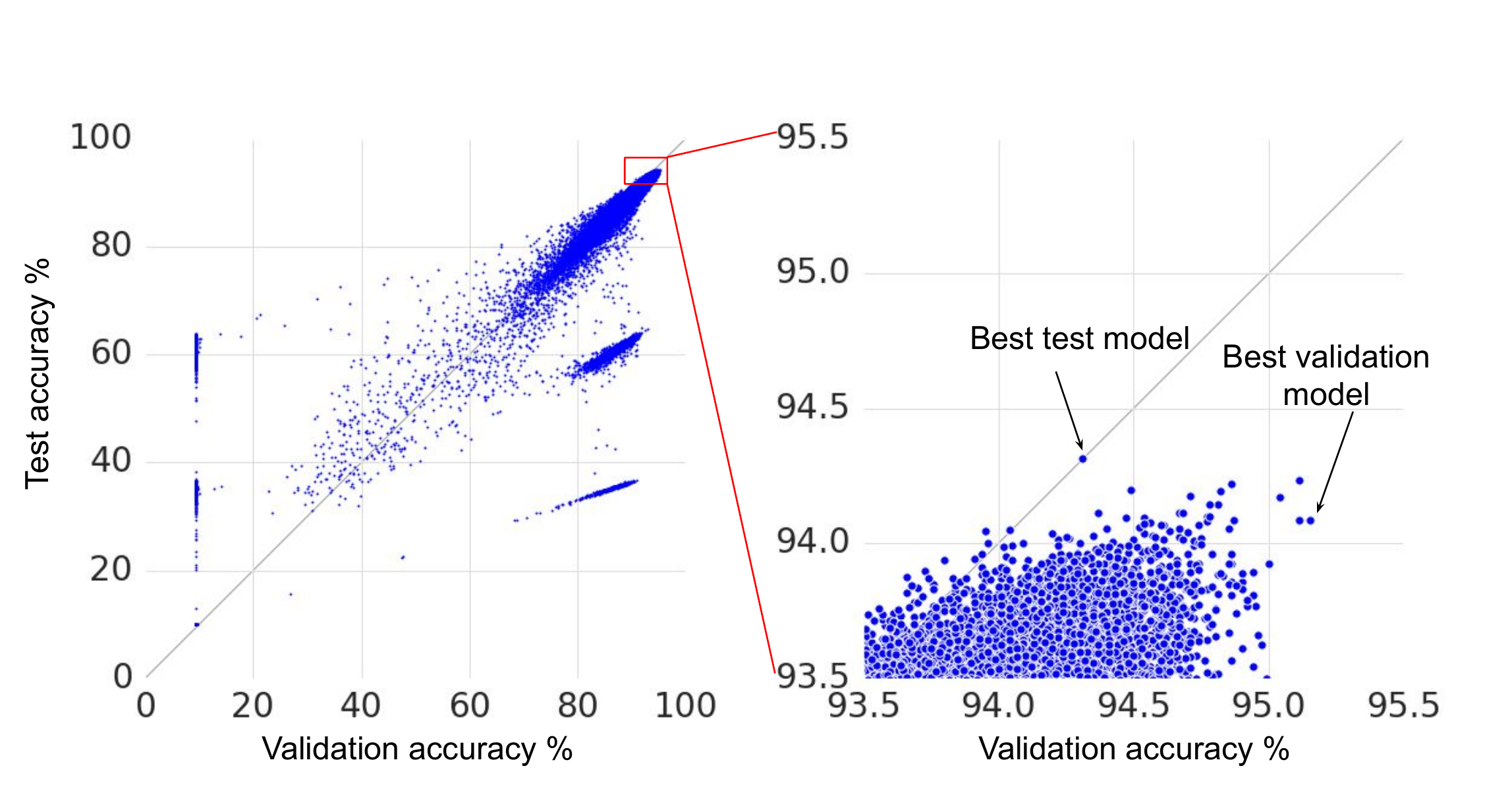}
\vspace{-12pt}
\end{center}
  \caption{(Left) Validation vs. test accuracy in NASBench-101. (Right) Zoomed in on the highly accurate region. Each model (point) is the validation accuracy from a single training run. Test accuracies are averaged over three runs. This plot demonstrates that even knowing the validation accuracy of every possible model is not sufficient to predict which model will perform best on the \emph{test} set.}
\label{fig:oracle_val_vs_test}
\end{figure}

\subsubsection{Oracle: an upper bound baseline}
Under the assumption of infinite compute, a traditional machine learning approach would be to train and validate all possible architectures to select the best one. We refer to this baseline as the ``oracle'' method. Figure~\ref{fig:oracle_val_vs_test} plots the validation versus the test accuracy for all models. The model that the oracle method would select based on the validation accuracy of  $95.15\%$ has a test set accuracy of $94.08\%$. This means that \textbf{the oracle does not select the model with the highest test set accuracy.} The global optimum on the test set is $94.32\%$. However, since this model cannot be found using extensive validation, one should not expect this model to be found using a well-performing architecture search algorithm. A more reasonable goal is to reliably select a model that has similar quality to the one selected by the oracle.
Furthermore, it is important to realize that \textbf{even an oracle approach has variance.} We have three training runs for each model, which allows us to run multiple variations of the ``oracle''. This simulates the impact of random variations on the final result. Averaged over $100$ oracle experiments, where in each experiment we randomly select one of 3 validation results, the best validation accuracy has a mean $95.13\%$ and a standard deviation $0.03\%$. 
The test accuracy has a mean of $94.18\%$ and a standard derviation $0.07\%$. 
% [95.14878112 94.05413593]
% [0.02881879 0.16165476]

\subsubsection{Random search: a lower bound baseline}
Recently, Li~\textit{et al.}~\cite{li2019random} questioned whether architecture search methods actually outperform random search. Because this depends heavily on the search space and Li~\textit{et al.}~\cite{li2019random} did not investigate the NASBench-101 search space we need to check this ourselves. Therefore we replicate the random baseline from NASBench-101 by sampling architectures without replacement. After training, we pick the architecture with the highest validation accuracy and report its result on the test set. Here we observe that even when we train and validate 2000 models, the gap to the oracle is large (Figure~\ref{fig:test_result_predictor}). For random search the average test accuracy is $93.66\%$ compared to $94.18\%$ for the oracle. This implies that 
\textbf{there is a large margin for improvement over random search.}
 Moreover, the variance is quite high, with a standard deviation of $0.25\%$. 
 
\subsubsection{Regularized evolution: a state of the art baseline}
In the original NASBench-101 publication, Regularized Evolution~\cite{real2019regularized} was the best performing method. We replicated those experiments using the open source code and their hyper-parameter settings (available in the supplementary material). %where the population size is set to $100$, the sampling size from the population is set to $10$, the mutation probabilities of edges and nodes are $\frac{1}{14}$ and $\frac{1}{10}$, respectively. 
%\footnote{\url{https://colab.research.google.com/github/google-research/nasbench/blob/master/NASBench.ipynb}}
\textbf{Regularized evolution is significantly better than random}
as shown in Figure~\ref{fig:test_result_predictor}. However even after $2000$ models are trained, it is still clearly worse than the oracle (on average) with an accuracy of $93.97\%$ and a standard deviation of $0.26\%$.

\subsubsection{Neural Predictor}

\begin{figure}
\begin{center}
\includegraphics[width=\columnwidth]{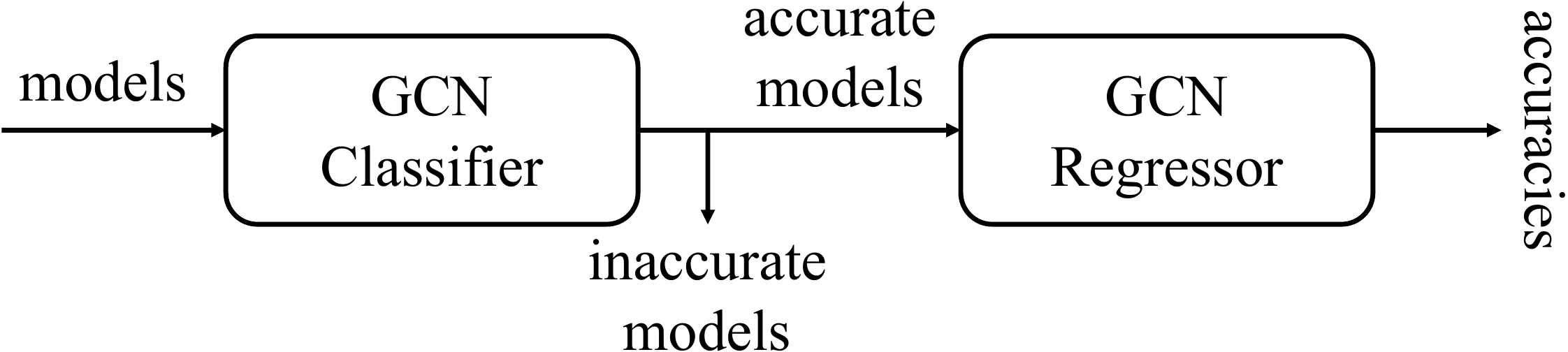}
\end{center}
  \caption{Neural Predictor on NASBench-101. It is a cascade of a classifier and a regressor. The classifier filters out inaccurate models and the regressor predicts accuracies of accurate models.}
\label{fig:cascade-classifier-regressor}
\end{figure}

\begin{figure}
\begin{center}
\includegraphics[width=\columnwidth]{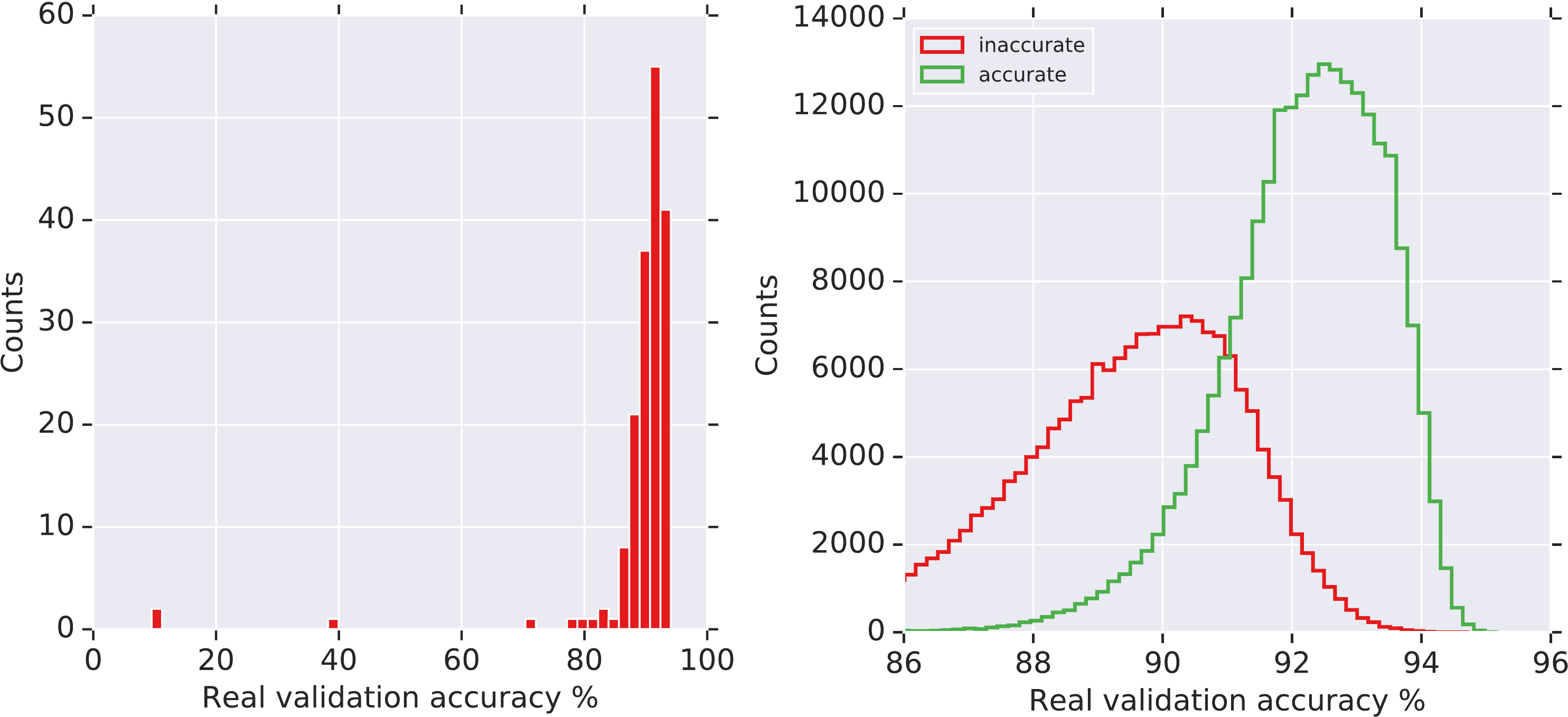}
\end{center}
  \caption{Binary classifier to filter out inaccurate models in NASBench-101. In this example, $172$ models (left) are sampled to cross validate the classifier. The right figure shows the performance of the classifier on unseen test models.}
\label{fig:classifier-eval}
\end{figure}

\begin{figure*}
\begin{center}
\includegraphics[width=.95\textwidth]{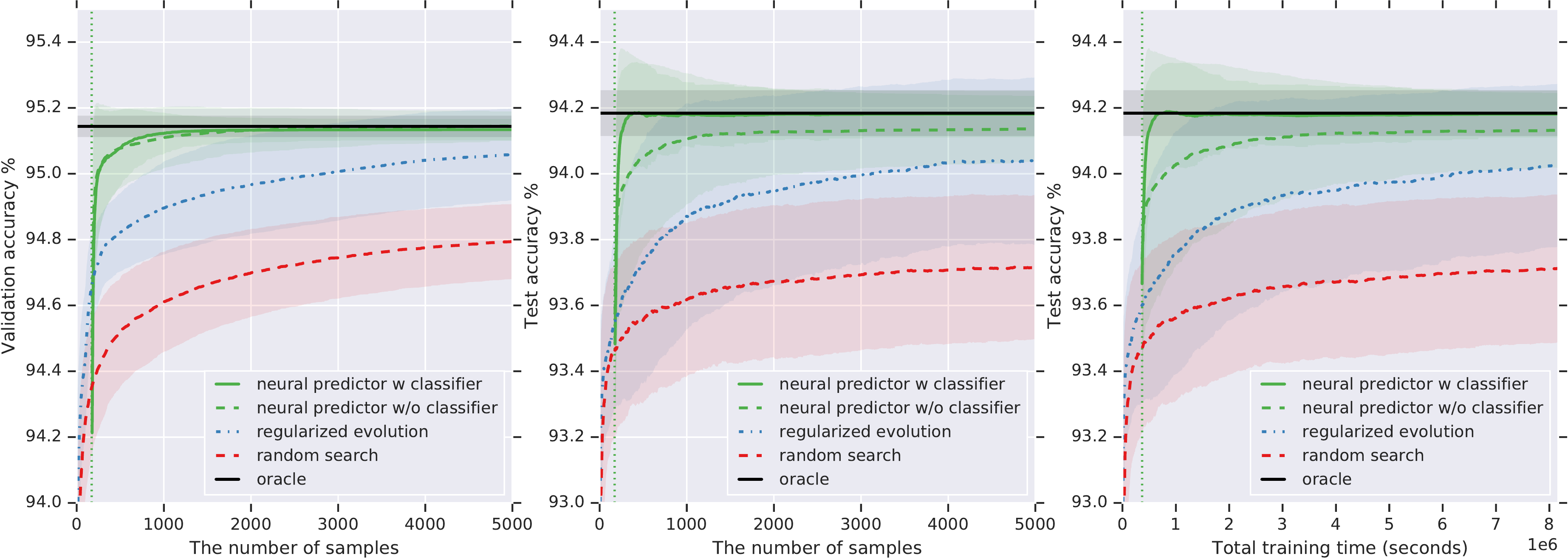}
\end{center}
\vspace{-12pt}
   \caption{The comparison of search efficiency among oracle, random search, Regularized Evolution and our Neural Predictor (with and without 2 stage regressor).
   All experiments are averaged over $600$ runs.
   The x-axis represents the total compute budget $N+K$.
   The vertical dotted line is at $N=172$ and represents the number of samples (or total training time) used to build our Neural Predictor. From this line on we start from $K=1$ and increase it as we use more architectures for final validation.
   The shaded region indicates standard deviation of each search method.
   }
\label{fig:test_result_predictor}
\end{figure*}

\begin{figure*}
\begin{center}
\includegraphics[width=0.95\textwidth]{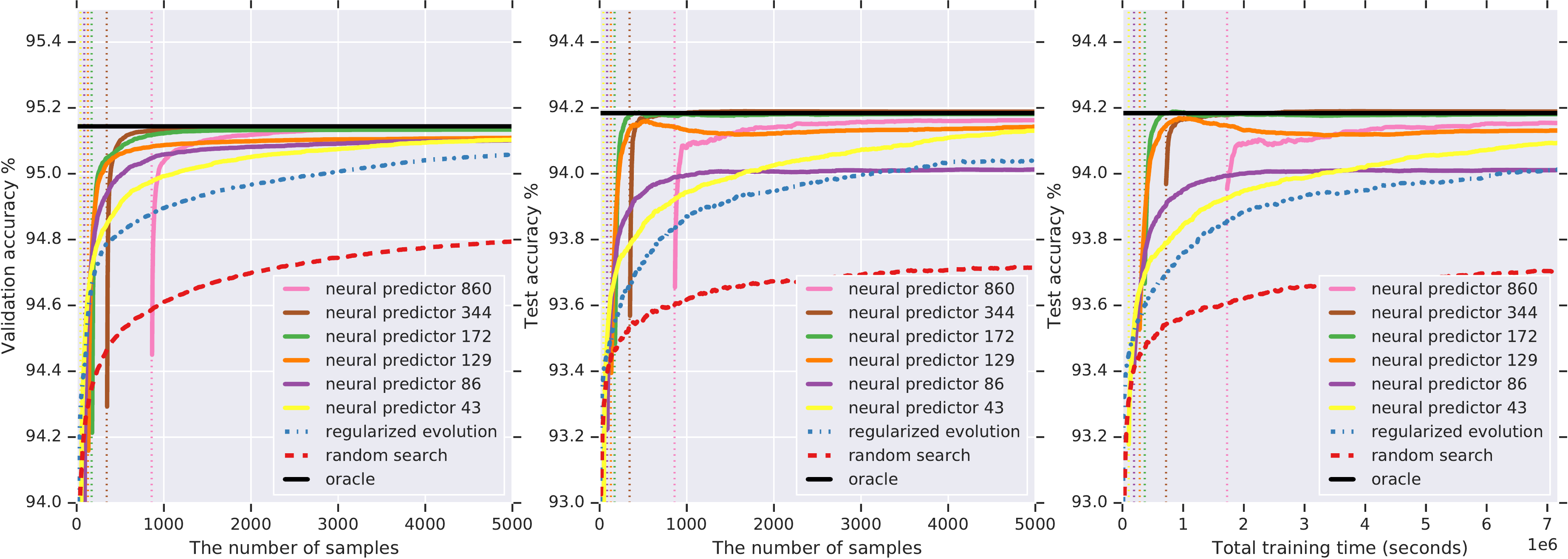}
\end{center}
\vspace{-12pt}
   \caption{Analysis of the trade-off between $N$ training samples vs $K$ final validation samples in the neural predictor. The x-axis is the total compute budget $N+K$. The vertical lines indicate different choices for $N$, the number of training samples and the point where we start validating $K$ models.
   All experiments are averaged over $600$ runs.}
\label{fig:comparison}
\end{figure*}

Having set our baselines, we now describe the precise Neural Predictor setup and evaluation.

\textbf{Setting up the GCN}
The graph representation of a model is a DAG with up to $7$ nodes. Each node is represented by an one-hot code of ``[\texttt{input}, \texttt{conv1x1}, \texttt{conv3x3}, \texttt{max-pool}, \texttt{output}]''.
 The GCN has three Graph Convolutional layers with the constant node representation size $D$ and one hidden fully-connected layer with output size $128$. Finally, the accuracy we need to predict is limited to a finite range. While it is not that common for regression, we can force the network to make predictions in this finite range by using a sigmoid at the output layer. Specifically, we use a sigmoid function that is scaled and shifted such that its output accuracy is always between $10\%$ and $100\%$. 

All hyper-parameters for the predictor are first optimized using cross-validation where $\frac{1}{3}N$ samples were used for validation. After setting the hyper-parameters, we use all $N$ samples to train the final predictor. At this point
we heuristically increase the node representation size $D$ such that the number of parameters in the Neural Predictor is also $1.5 \times$ larger.
Specific $N$ and $D$ values and other training details are in the supplementary material.

\textbf{A two stage predictor}
Looking at a small dataset of $N=172$\footnote{In our implementation, we split the NASBench-101 dataset to $10,000$ shards and each shard has $43$ samples. The $N=172$ comes from a random $4$ shards.} models in Figure~\ref{fig:classifier-eval} (left) during cross-validation, we realized that for NASBench-101 a two stage predictor is needed. The NasBench-101 dataset contains many models that are not stable during training or perform very poorly (e.g. a model with only pooling operations). 

The two stage predictor, shown in Figure~\ref{fig:cascade-classifier-regressor}, filters obviously bad models first by predicting whether each model will achieve an accuracy above 91\%. This allows the the second stage to focus on a narrower accuracy range, which makes it more reliable.

Both stages share the same GCN architecture but have different output layers. A classifier trained on these $N=172$ models has a low False Negative Rate as shown in Figure~\ref{fig:classifier-eval} (right). This implies that the classifier will filter out very few actually good models. 

If we only use a single stage, the MSE for the validation accuracy is 1.95 (averaged over 10 random splits). By introducing the filtering stage this reduces to 0.66. In our final results in in Figure~\ref{fig:test_result_predictor} we will also include an evaluation of the predictor without the filtering stage.

\textbf{Results using $N=172$} (or $0.04\%$ of the search space) for training are shown in Figure~\ref{fig:test_result_predictor}. We used $N=172$ models to train the predictor. 
Then we vary $K$, the number of architectures with the highest predicted accuracies to be trained and validated to select the best one. Therefore, ``the number of samples'' in the figure equals $N+K$ for Neural Predictor.
In Figure~\ref{fig:test_result_predictor} (left), our Neural Predictor significantly outperforms Regularized Evolution in terms of sample efficiency. On the test set, the mean validation accuracy is comparable to that of the oracle after about 1000 samples.
The sample efficiency in validation accuracy transfers well to test accuracy in terms of both the total number of trained models in Figure~\ref{fig:test_result_predictor} (middle) and wall-clock time in Figure~\ref{fig:test_result_predictor} (right).
After $5000$ samples, Regularized Evolution reaches validation and test accuracies of $95.06\%$ and $94.04\%$ respectively; our predictor can reach the same validation $12.40 \times$ faster and the same test $22.83 \times$ faster.
%and Here the results indicate that the proposed approach is at least 10 times more efficient than Regularized Evolution.
Another advantage we observe is that Neural Predictor has small search variance.

%The variance of the neural predictor decreases as the number of models $K$ for final evaluation increases. This might seem counter-intuitive but can be explained easily. As long as the predictor has high recall for a certain $K$, most of the truly good models will be selected for final evaluation. hence, if most of the good models are evaluated, the behavior will be similar to that of the oracle. This implies of course that the tradeoff between $N$ and $K$ is very important, which we analyze next. 

\textbf{$N$ vs $K$ and ablation study.}~
We next consider the problem of choosing an optimal value of $N$ when the total number of models we're permitted to train, $N + K$, is fixed. %; that is, given the total number of models ($N+K$) we can train, what is an optimal $N$?
Figure~\ref{fig:comparison} summarizes our study on $N$. A Neural Predictor underperforms with a very small $N$ ($43$ or $86$), as it cannot predict accurately enough which models are interesting to evaluate. 
In Figure~\ref{fig:oracle_val_vs_test} we have shown that some models are higher ranked according to  validation accuracies than test accuracies. This can cause the test accuracy to degrade as we increase $K$.

Finally, we consider the case where $N$ is large (e.g., $N=860$) but $K$ is small. In this case we clearly see that the increase in quality of the GCN cannot compensate for the decrease in evaluation budget.
% We also observe that Neural Predictor can achieve test accuracy higher than the ``oracle'' bound. This is also normal because of the same reason.
% The ``oracle'' bound is the bound of validation accuracy instead of test accuracy.
% The ``oracle'' lines in Figure~\ref{fig:test_result_predictor} (middle and right) are just test accuracy of the model with oracle validation accuracy in Figure~\ref{fig:test_result_predictor} (left).
% A model with a validation accuracy lower than its bound can achieve a test accuracy higher than the bound model.

% \begin{figure*}
% \begin{center}
% \fbox{\rule{0pt}{2in} \rule{.9\linewidth}{0pt}}
% \end{center}
%   \caption{In panel A we show the correlation between the validation and the test accuracies on the NASBench dataset. In panel B \ww{[It will be complicated since I used a classifier to filter low accurate models.]} we show the correlation between the validation and the neural predictor on NASBench. in panel C we show the correlation between the predictor and the test accuracy. In all plots we sampled the same subset of the search space. These were not used for training the predictor. {\color{red} An option would be to show the plot with the training data marked in a special color, another option would be to not do the correlation between predictor and test, but do the predictor at multiple values for the amount of training data.}}
% \label{fig:validation_vs_test_vs_predictor}
% \end{figure*}

%%%%%%%%%%%%%%%%%%%%%%%%%%%%%%%%%%%%%%%%%%%%%%%%%%%%%%%%%%%%%%%%%%%%%%%%%%%%%%%%%%%%%%%%%%%%%%%%%%%%%%%%%%%%%%%%%%%%%%%%%%%%%%%%%%%

\subsection{ImageNet Experiments}

\begin{figure*}
\begin{center}
\includegraphics[width=0.75\textwidth]{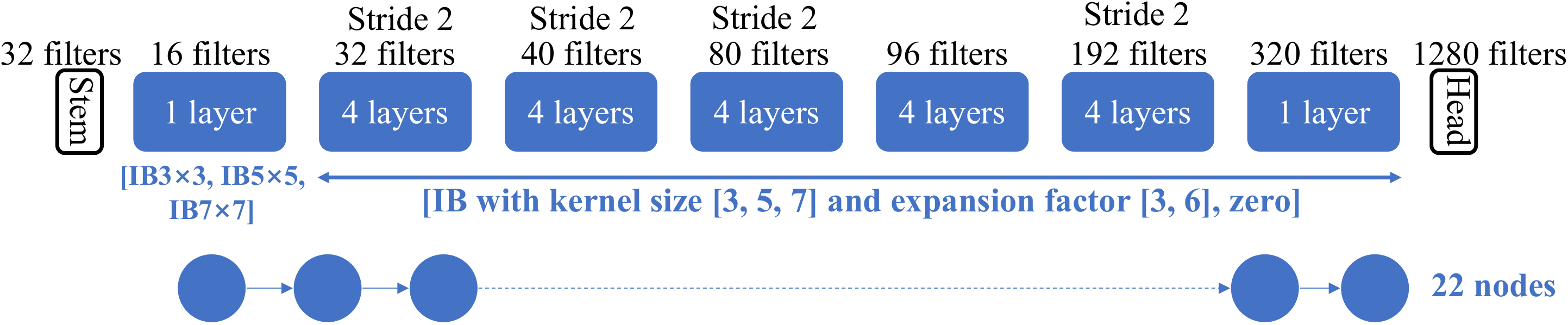}
\end{center}
\vspace{-12pt}
   \caption{The search space of ProxylessNAS. Only convolutional layers in blue are searched. The optional operations in each layer are $6$ types of Inverted Bottleneck (IB)~\cite{sandler2018mobilenetv2} (with a kernel size $3 \times 3$, $5 \times 5$ or $7 \times 7$ and an expansion factor of $3$ or $6$) and one zero operation (which outputs zeros for layer skipping purpose).
   The expansion factor in the first block is fixed as $1$, and the zero operation is forbidden in the first layer of every block.
   The search space size is $3 * 6 ^ {6} * 7 ^ {15} \approx 6.64 \times 10 ^ {17} $.}
\label{fig:proxylessnas_space}
\end{figure*}

\begin{figure*}
\begin{center}
\includegraphics[width=0.9\textwidth]{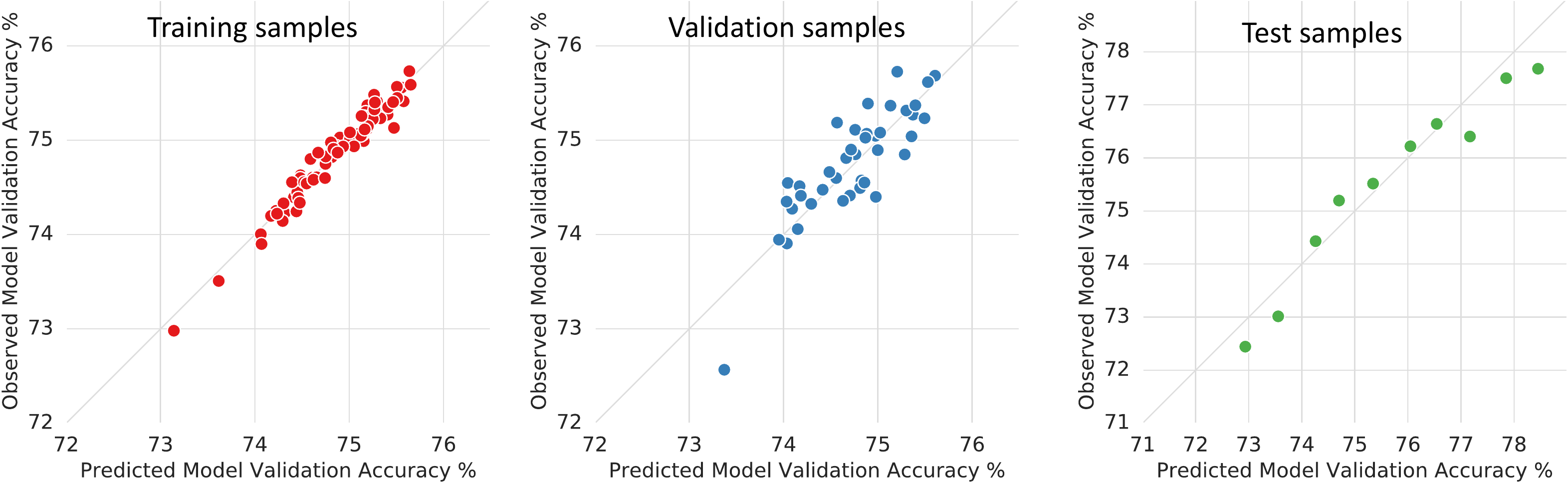}
\end{center}
\vspace{-12pt}
   \caption{The performance of Neural Predictor on training, validation and test samples. %\gb{The model accuracies on the test samples range between 72.5\% and 77.5\% but on the validation samples, they range between 72.5\% and 76\% (and are mostly between 74\% and 76\%). Why the discrepancy?}}
   }
\label{fig:imagenet_predictor_perf}
\end{figure*}

\begin{figure*}
\begin{center}
\includegraphics[width=0.95\textwidth]{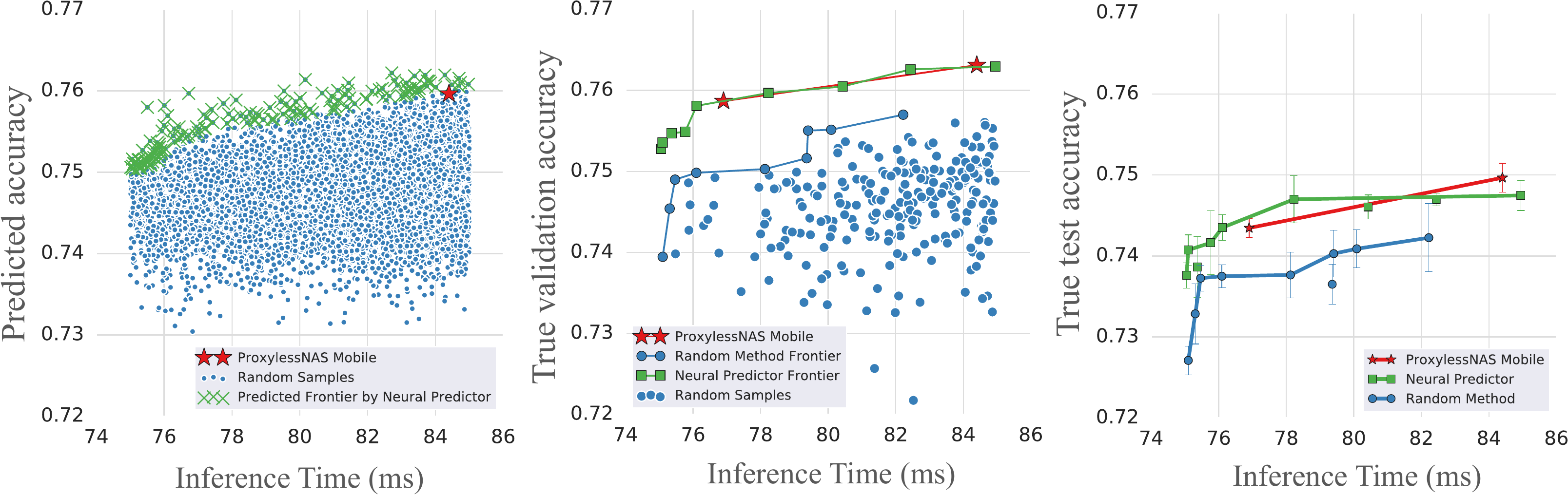}
\end{center}
\vspace{-12pt}
    \caption{Comparison between random search, ProxylessNAS and Neural Predictor.
    \textbf{Left:} Frontier models predicted by the Neural Predictor.
    \textbf{Middle:} Validation accuracy of each found frontier models.
    \textbf{Right:} Test accuracy of each found frontier model. (Each test accuracy is averaged over $5$ training runs under different initial weight values. Error bars with $95\%$ confidence interval are also plotted in the figure.)}
\label{fig:image_compare}
\end{figure*}

While the NASBench-101 dataset allows us to look at the behavior in a well controlled environment, it does not allow us to evaluate whether the approach generalizes to larger scale problems. It also does not address the issue of finding high \textit{quality} inference time constrained models.
Therefore, to demonstrate that our approach is more widely applicable we look at this use case in our second set of experiments on ImageNet~\cite{deng2009imagenet} with the ProxylessNAS search space \cite{cai2018proxylessnas}. We will compare our results to a random baseline and our own reproduction of the ProxylessNAS search. 
In this search space, the goal is to find a good model that has an inference time around 84 ms on a Pixel-1 phone.

\textbf{Search space}
The ProxylessNAS search space is illustrated in Figure~\ref{fig:proxylessnas_space}. It does not have the cell-based structure from NASBench-101; it instead requires independent choices for the individual layers. The layers are divided up into blocks, each of which has its own fixed resolution and a fixed number of output filters. We search over which layers to skip and what operations to use in each layer. (The first layer of a block is always present.) There are approximately $6.64 \times 10 ^ {17} $ models in the search space. Because this search space is so large, we cannot generate the oracle baseline; we must instead rely on the random search and ProxylessNAS re-implementations we discuss next.

\subsubsection{Baselines}
\textbf{The random search baseline } samples $256$ models with inference times between $75$ms and $85$ms. All these models are trained for 90 epochs.% and validated.
Then, we look at which models are Pareto optimal (i.e. have good tradeoffs between inference time and validation quality). 
All Pareto optimal models were then trained for 360 epochs
and evaluated on the test set. The results are shown in Figure~\ref{fig:image_compare}. More implementation details are in the supplementary material.

\textbf{ProxylessNAS~\cite{cai2018proxylessnas}} is an efficient architecture search algorithm based on weight sharing and reinforcement learning (RL). It trains a large neural network where different paths can be switched on or off to mimic specific architectures in the search space. The RL controller assumes that for a single network all decisions can be made independently (i.e. the probability distribution over architectures is factorized). To train the shared weights of the large model, we repeat the process of (i) sampling an architecture from the RL controller and (ii) training it for a single step. To update the RL controller, another batch is sampled. This time the batch is evaluated on the validation set, and this result is used in combination with additional information (i.e. the latency) to compute a reward used to update the RL controller. 

In the original publication, ProxylessNAS reports an accuracy of 74.6\% for their best model; our reproduction of that model achieves 74.9\% accuracy. We also re-implemented the search algorithm itself.  By repeating the search 5 times we obtain an average accuracy of 75.0\% and a variance of 0.1.% Because the original ProxylessNAS implementation is not public it is unclear whether the minor difference to the published model is due to the variance of the search, the initialization or another factor.
Since these results are close together we consider this sufficiently good as a basis for comparison for the neural predictor. 

\subsubsection{Neural predictor}
Overall, we use the same basic pipeline as in the NASBench-101 experiments. However, because the models in the ProxylessNAS search space are much more stable than those in NASBench-101 we only need a single stage predictor. To transfer from NASBench-101 to the ProxylessNAS search space, all we have to do is to modify the node representation. Because the ProxylessNAS search space is just a linear graph as shown in Figure~\ref{fig:proxylessnas_space} (bottom), we can modify the node representation at the input to be nothing more than a one hot vector with length $7$ as this allows us to describe all architectures.

\textbf{Training and validating the neural predictor}
To build the neural predictor we randomly sample $119$ models; $79$ samples are used for training and $40$ samples are for validation.
To find the GCN's hyperparameters we average validation MSE scores over $10$ random  training and validation splits. Based on this we select a GCN with $18$ Graph Convolutional layers with node representation size $96$, and with two fully-connected layers with hidden sizes $512$ and $128$ on top of the mean node representations in the last Graph Convolutional layer. After all hyper-parameters are finalized, we train our GCN with all $119$ samples.

Our validation also showed that for ImageNet experiments, no classifier is needed to filter inaccurate models. This is because the model accuracy is within a relatively small range as  as shown in Figure~\ref{fig:imagenet_predictor_perf} (left and middle).
Our final settings for the Neural Predictor achieved on MSE $0.109 \pm 0.028$ averaged over $10$ validation runs. 
Figure~\ref{fig:imagenet_predictor_perf} shows an example of the correlation between true accuracy and predicted accuracy for training samples (left) and validation samples (middle).
For validation samples, the Kendall rank correlation coefficient is $0.649$ and the Coefficient of Determination ($R^2$ score) is $0.648895$.
%\gb{If the prediction errors on the validation set were normally distributed, I would've expected the difference between the predicted and observed accuracies to be less than $3 * \text{MSE} \approx 0.3$ in more than 99\% of cases. But in Figure 9, it seems like it's relatively common for them to differ by 0.5. What am I missing?}

\textbf{Looking at the predictive performance of the predictor.}
In Figure~\ref{fig:imagenet_predictor_perf} (right), we test the generalization of our Neural Predictor to unseen test architectures.
We first randomly sample $100K$ models from the ProxylessNAS search space without inference time constraint and predict their accuracies. We then pick the model with minimum predicted accuracy ($72.94\%$), the model with maximum predicted accuracy ($78.45\%$), and $8$ additional models which are evenly spaced between those two endpoints. We train those $10$ models to obtain their true accuracies.
In Figure~\ref{fig:imagenet_predictor_perf} (right), the Kendall rank correlation coefficient is $0.956$ and the Coefficient of Determination ($R^2$ score) is $0.929$. % $0.928737$.
More interesting, although our training dataset never observed models with accuracies higher than $76\%$, our Neural Predictor can still successfully predict the $5$ models with accuracies higher than $76\%$. This demonstrates the generalization of our Neural Predictor to unseen data.

\textbf{Finding high quality mobile sized models}
We now use the predictor to select frontier models with good trade-offs between accuracies and inference times.
We randomly sample $N=\text{112,000}$ models with inference times between $75$ms and $85$ms, and predict  their accuracies as shown in Figure~\ref{fig:image_compare} (left).
As a sanity check, we also predict the quality of the ProxylessNAS model. The predicted validation accuracy is $76.0\%$ which is close to its true validation accuracy $76.3\%$.

The next step is selecting $K$ Pareto optimal models. However, because the predictor can make mistakes, we need a soft version of Pareto optimality. 
To do so, we sort the models based on increasing inference time. In the regular definition, a model is Pareto optimal if no faster model has higher quality.
In our setup, we define a model as "soft-Pareto optimal" when the predicted accuracy is higher than the minimum of the previous $J$ models. In our experiments we set $J=6$. This leaves us with $137$ promising models in green in Figure~\ref{fig:image_compare} (left).

All $k=137$ models are then trained and validated. This allows us to obtain a traditional Pareto frontier as shown in Figure~\ref{fig:image_compare} (middle).
The architectures of those true frontier models are included in the supplementary material.
%%% DO NOT DELETE!!! DO NOT DELETE!!! DO NOT DELETE!!! DO NOT DELETE!!! DO NOT DELETE!!!
% [(0.7376040101051331, 75.05052185058594, (0, 0, 6, 0, 0, 0, 0, 0, 6, 1, 4, 6, 2, 4, 0, 1, 5, 0, 2, 6, 2, 3)), (0.7407359957695008, 75.09565734863281, (0, 0, 6, 0, 6, 1, 2, 0, 6, 3, 0, 1, 5, 2, 0, 0, 1, 4, 0, 6, 5, 3)), (0.738647997379303, 75.36453247070312, (0, 1, 6, 6, 6, 0, 4, 6, 6, 4, 6, 6, 1, 5, 5, 1, 3, 1, 5, 2, 2, 3)), (0.7416280031204223, 75.76422119140625, (0, 0, 6, 6, 6, 2, 0, 2, 3, 4, 6, 1, 6, 0, 1, 1, 1, 5, 4, 2, 3, 3)), (0.7434999942779541, 76.10467529296875, (0, 0, 6, 0, 6, 2, 3, 6, 0, 4, 3, 4, 5, 3, 0, 6, 0, 1, 0, 2, 2, 3)), (0.7470040082931518, 78.22792053222656, (0, 0, 0, 6, 6, 1, 0, 6, 0, 5, 0, 2, 1, 4, 0, 2, 2, 5, 5, 0, 2, 3)), (0.7460600018501282, 80.42455291748047, (0, 0, 6, 0, 6, 2, 0, 0, 2, 1, 0, 0, 6, 5, 0, 1, 3, 5, 5, 2, 2, 3)), (0.7469719886779785, 82.44024658203125, (0, 0, 6, 6, 0, 2, 0, 0, 4, 5, 4, 6, 2, 4, 3, 3, 6, 5, 1, 2, 2, 3)), (0.7474720120429993, 84.94596862792969, (0, 0, 0, 0, 6, 1, 3, 1, 6, 4, 2, 1, 0, 2, 6, 0, 5, 4, 2, 2, 5, 3))]
\textbf{The Neural Predictor outperforms the random baseline and is comparable to ProxylessNAS.}
Recall that the random baseline trained $256$ models. This is the same number of models we trained in total for our method ($N=119$ models for training the predictor and $K=137$ models selected for final validation). 
For test set comparisons we train our models and our reproduction of ProxylessNAS for 360 epochs. The results are shown in Figure~\ref{fig:image_compare} (right).%\gb{The preceding sentence could also benefit from some rewording.} 
 Now we observe that the gap between the Pareto frontier of the neural predictor and the random baseline is stable on unseen data.
To  obtain a frontier for ProxylessNAS, one could run multiple searches to reduce/increase the inference time. 
We opt to reduce the number of filters in each layer to $0.92 \times$ to obtain a faster model. 
The results show that the Neural Predictor and ProxylessNAS perform comparably.
\section{Discussion and Related Work}
 The effectiveness of NAS vs random search has been questioned recently \cite{li2019random}.
 On NASBench-101 the Neural Predictor and Regularized Evolution are clearly better than random search. We also observed that the Neural Predictor is about $22.83$ times more efficient than Regularized Evolution on NASBench-101.
 
In the ProxylessNAS search space we saw that the Neural Predictor and ProxylessNAS produce models of similar quality. We believe that implementing and using the Neural Predictor is more straightforward but running a single ProxylessNAS search requires fewer resources.
Training all $N+K=256$ models for the entire Neural Predictor experiment took 47.5 times as much compute as a single ProxylessNAS search.
In practice the gap is actually much smaller because optimizing the hyper-parameters of the Neural Predictor has negligible cost but trying out a new hyper-parameter configuration for ProxylessNAS requires a full search. On top of that, the Neural Predictor is more effective at targeting different latency targets than ProxylessNAS, which needs a search per target. Furthermore, in the ideal setting where we can completely parallelize model training for the Neural Predictor ($N=119$ models for training in parallel followed by $K=136$ for validation in parallel) it would finish in half the time of a ProxylessNAS search. Based on this analysis we believe that the Neural Predictor and ProxylessNAS are complementary. Choosing which method to use will depend on the complexity of implementing the search space (with weight sharing) and the computational resources available.

Finally, we want to point out that this approach is not the first to use a regression model in a NAS setup. However, it is by far the most straightforward one. Other approaches use more complicated setups~\cite{liu2018progressive, baker2017accelerating}, propose creative ways to back-propagate through network architectures~\cite{luo2018neural}, or apply Bayesian optimization~\cite{dai2019chamnet} or complicated weight sharing~\cite{bender2018understanding} instead of a simple regression model.

\section{Acknowledgements}
We would like to thank Chris Ying, Ken Caluwaerts, Esteban Real, Jon Shlens and Quoc Le for valuable input and discussions.

{\small
\bibliographystyle{ieee_fullname}
\bibliography{main}
}
\cleardoublepage

\begin{table}[b]
\begin{center}
 \begin{tabular}{|c c| |c c|} 
 \hline
 $N$ & $D$ & $N$ & $D$  \\
 $43$ & $48$ & $172$ & $144$ \\ 
 $86$ & $72$ & $334$ & $210$ \\ 
 $129$ & $96$ & $860$ & $320$ \\
 \hline
\end{tabular}
\caption{Node representation size $D$ under $N$}
\label{tab:n_d}
\end{center}
\end{table}

\begin{figure*}
\begin{center}
\includegraphics[width=1.\textwidth]{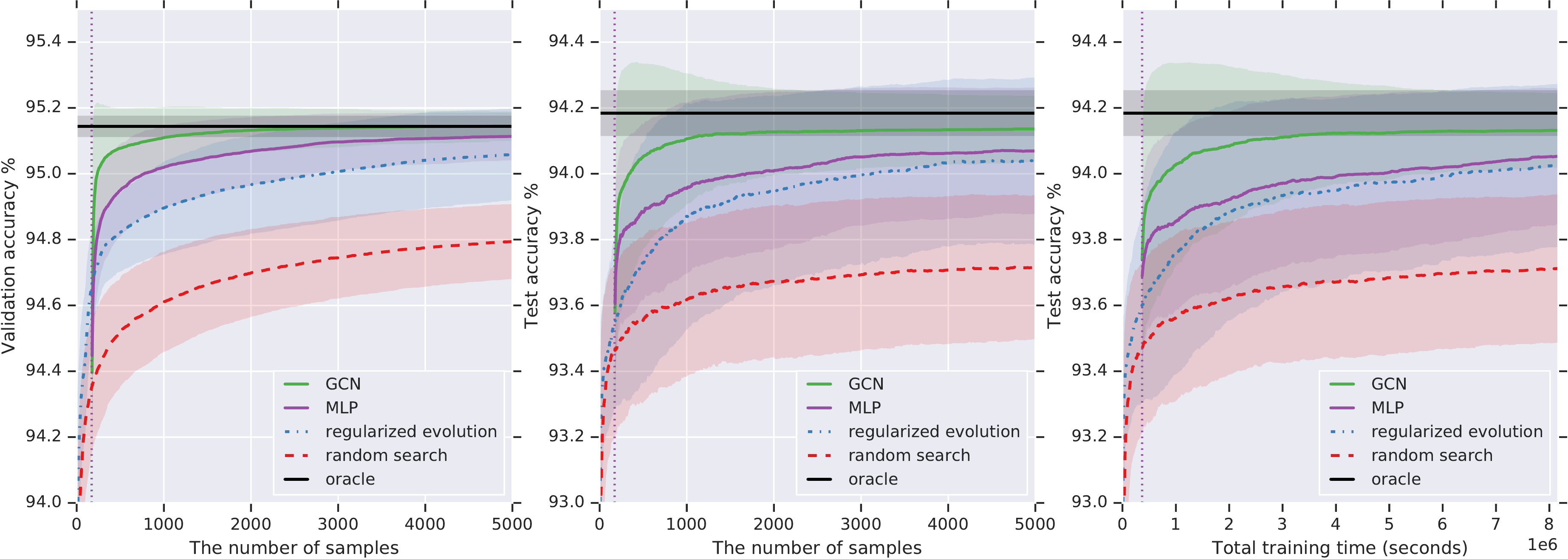}
\end{center}
\vspace{-12pt}
   \caption{The ablation study of our Neural Predictor under different architectures.
   Experiments are performed in NASBench-101.
   In this study, a one stage predictor without a classifier is used.
   All methods are averaged over $600$ experiments.
   The shaded region indicates standard deviation of each search method.
   The x-axis represents the total compute budget $N+K$.
   The vertical dotted line is at $N=172$ and represents the number of samples (or total training time) used to build our Neural Predictor. From this line on we start from $K=1$ and increase it as we use more architectures for final validation.
   }
\label{fig:gcn_vs_mlp}
\end{figure*}

\begin{table*}
\begin{center}
 \begin{tabular}{c|c|c} 
 \hline
 Inference time & Top-1 test accuracy & Architecture  \\
 \hline
 $ 75.05 $ms & $ 73.76 \pm 0.08 \% $ & $ (0,0,6,0,0,0,0,0,6,1,4,6,2,4,0,1,5,0,2,6,2,3) $ \\ 
 $ 75.10 $ms & $ 74.07 \pm 0.09 \% $ & $ (0,0,6,0,6,1,2,0,6,3,0,1,5,2,0,0,1,4,0,6,5,3) $\\ 
 $ 75.36 $ms & $ 73.86 \pm 0.19 \% $ & $ (0,1,6,6,6,0,4,6,6,4,6,6,1,5,5,1,3,1,5,2,2,3) $\\ 
 $ 75.76 $ms & $ 74.16 \pm 0.20 \% $ & $ (0,0,6,6,6,2,0,2,3,4,6,1,6,0,1,1,1,5,4,2,3,3) $\\ 
 $ 76.10 $ms & $ 74.35 \pm 0.08 \% $ & $ (0,0,6,0,6,2,3,6,0,4,3,4,5,3,0,6,0,1,0,2,2,3) $\\ 
 $ 78.23 $ms & $ 74.70 \pm 0.15 \% $ & $ (0,0,0,6,6,1,0,6,0,5,0,2,1,4,0,2,2,5,5,0,2,3) $\\ 
 $ 80.42 $ms & $ 74.61 \pm 0.07 \% $ & $ (0,0,6,0,6,2,0,0,2,1,0,0,6,5,0,1,3,5,5,2,2,3) $\\ 
 $ 82.44 $ms & $ 74.70 \pm 0.04 \% $ & $ (0,0,6,6,0,2,0,0,4,5,4,6,2,4,3,3,6,5,1,2,2,3) $\\ 
 $ 84.95 $ms & $ 74.75 \pm 0.09 \% $ & $ (0,0,0,0,6,1,3,1,6,4,2,1,0,2,6,0,5,4,2,2,5,3) $\\ 
 \hline
\end{tabular}
\caption{Frontier Architectures in Figure 10 (right) in Section 3.2.2. An architecture is represented by the indices of operations in all $22$ layers. The mapping between indices and operations are listed in Figure 8 in Section 3.2.
In the first block (layer), an index selects [IB3x3, IB5x5, IB7x7] with a fixed expansion factor $1$.
In other layers, an index selects [IB3x3-3, IB5x5-3, IB7x7-3, IB3x3-6, IB5x5-6, IB7x7-6, zero], where a suffix ``-$M$'' denotes an expansion factor $M$.
}
\label{tab:archs}
\end{center}
\end{table*}

\section{Supplementary Material}
\subsection{Ablation Study of Neural Predictor Architectures}
Figure~\ref{fig:gcn_vs_mlp} includes ablation study of different architectures for the Neural Predictor on NASBench-101.
We compared Graph Convolutional Networks (GCN) and Multi-layer Perceptrons (MLP) in the figure.
To generate inputs for a MLP, we simply concatenate the one-hot codes of node operations with the upper triangle of the adjacency matrix.
From the figure, we can see such a simple MLP can outperform state-of-the-art Regularized Evolution; more importantly, the 
GCN that we selected achieves the best.
We also tried Convolutional Neural Networks (CNN) but completely failed with a performance near to random search.
During our development, we also proposed a data augmentation to improve the performance of MLP and CNN. In this augmentation, we randomly permute the order of nodes to generate new inputs online.
However, we needed to perform the permutation during validation; otherwise, the validation data distribution is different from training data distribution.
More importantly, GCN encodes the inductive bias that the prediction should be permutation invariant. Therefore, GCN is our final decision.

\subsection{Reproduction of Regularized Evolution~\cite{real2019regularized}}

We follow the NASBench-101 paper and their released code\footnote{\url{https://colab.research.google.com/github/google-research/nasbench/blob/master/NASBench.ipynb}} to reproduce Regularized Evolution. The population size is set to $100$, the sampling size from the population is set to $10$, and the mutation probabilities of edges and nodes are $\frac{1}{14}$ and $\frac{1}{10}$ respectively.

For reproduction purpose, we clarify two differences in this paper when plotting curves of ``the test accuracy versus training time spent'':
\begin{itemize}
    \item in the NASBench-101 paper, the test accuracy comes from a single training run, which leads to the use of a single validation accuracy as the signal for search. We instead report the mean test accuracy over three records. We use the mean because it is a quality expectation when the a discovered architecture is distributed and re-trained by different users, and it simulates a scenario where higher uncertainty exists. Moreover, it is the user case that we encountered in the ImageNet experiments.
    \item in Figure 6 and Figure 7 in Section 3.1, we plot the test accuracy averaged over $600$ experiments; while, in the NASBench-101 paper, the median test accuracy over $100$ experiments were plotted.
    % \gb{I had trouble following the preceding two sentences.}
\end{itemize}

\subsection{Implementation Details of Neural Predictor}
\subsubsection{NASBench-101 on Cifar-10}

\paragraph{The Architecture of Neural Predictor} starts with three bidirectional Graph Convolutional layers, whose node representations have the same size $D$.
%\gb{What's a consistent node representation?}
The node representations from the last Graph Convolutional layer are averaged to obtain a graph representation, which is followed by a fully-connected layer with hidden size $128$ and an output layer.

\paragraph{Training Hyper-parameters of the Neural Predictor} are cross validated. For the classifier in the two stage predictor, we use the Adam optimizer~\cite{kingma2014adam} with an initial learning rate $0.0002$, dropout rate $0.1$ and weight decay $0.001$. The learning rate is gradually decayed to zero by a cosine schedule~\cite{loshchilov2016sgdr}.
We train the classifier for $300$ epochs with a mini-batch size $10$.
The regressor in the two stage predictor uses the same hyper-parameters but an initial learning rate $0.0001$.

\paragraph{Node representation size $D$} under different training dataset size $N$ is listed in Table~\ref{tab:n_d}. % \gb{What is D? Can we include a reminder of what $D$ is here?}

\subsubsection{ProxylessNAS on ImageNet}
\paragraph{The Architecture of Neural Predictor} includes $18$ bidirectional Graph Convolutional layers.
The node representations from the last Graph Convolutional layer are averaged to obtain a graph representation, which is followed by two fully-connected layers with hidden sizes $512$ and $128$ and an output layer.
The predictor is one stage without a classifier.
All Graph Convolutional layers have a node representation size $96$.

\paragraph{The Training Hyper-parameters of Neural Predictor} are cross validated. We have $119$ samples\footnote{We trained $120$ models and one crashed, ending up with the odd number $119$}, $40$ of which are validation samples and $79$ are training samples.
Training hyper-parameters are validated by averaging over $10$ random splits of $119$ samples.
We use the Adam optimizer with an initial learning rate $0.001$ and weight decay $0.00001$.
The learning rate is gradually decayed to zero by a cosine schedule~\cite{loshchilov2016sgdr}.
We train for $300$ epochs with a mini-batch size $10$.

\paragraph{Training Hyper-Parameters of Image Classification Models on ImageNet}
The official ImageNet dataset consists of 1,281,167 training examples and 50,000 validation examples. Since the official test set for ImageNet was never publicly released, we follow the standard (although admittedly confusing) convention of using the 50,000-example validation set as our test set. We randomly partitioned the 1,281,167-image training set into 1024 shards, and used the final 40 shards -- or 50,046 examples -- as our validation set. For models which we planned to evaluate on our validation set, we excluded these 50,046 examples during model training. We did, however, use these examples for models we planned to evaluate on our test set.

When training image classification models, we used distributed synchronous SGD with four Cloud TPU v2 or v3 instances (i.e., 32 TPU cores) and a per-core batch size of 128. Models were optimized using RMSProp with momentum 0.9, decay 0.9, and epsilon 0.1. The learning rate was decayed according to a cosine schedule. Models were trained with batch normalization with epsilon 0.001 and momentum 0.99. Convolutional kernels were initialized with He initialization\footnote{The default TensorFlow implementation of He initialization has an issue which can cause it to overestimate the fan-in of depthwise convolutions by multiple orders of magnitude. We correct this issue in our implementation.} \cite{he2015delving} and bias variables were initialized to 0. We initialized the final fully connected layer of the network with mean 0 and stddev 0.01. We used an L2 regularization rate of $4 \times 10^{-5}$ for all convolutional kernels, but did not apply L2 regularization to the final fully connected layer. Models were trained on $224 \times 224$ input images with ResNet \cite{he2016deep} image preprocessing. Models were either trained for 90 or 360 epochs. We used a dropout rate of 0 (resp. 0.15) before the final fully connected layer when training models for 90 (resp. 360) epochs.
%\ww{Please help: dataset split for validation and test. Hyperparameters when training $90$ epochs and $360$ epochs.}

\subsection{Discovered Frontier Models on ImageNet}
Table~\ref{tab:archs} includes architectures at the frontier in Figure 10 (right) in Section 3.2.2.
Our predictor discovers architectures with cheap operations (with small kernel size and expansion factor) or the skip operation (i.e. zero by index $6$) in the early layers, and places diverse operations in later layers to make the trade-off.

%%% DO NOT DELETE!!! DO NOT DELETE!!! DO NOT DELETE!!! DO NOT DELETE!!! DO NOT DELETE!!!
% [(0.7376040101051331, 75.05052185058594, (0, 0, 6, 0, 0, 0, 0, 0, 6, 1, 4, 6, 2, 4, 0, 1, 5, 0, 2, 6, 2, 3)), (0.7407359957695008, 75.09565734863281, (0, 0, 6, 0, 6, 1, 2, 0, 6, 3, 0, 1, 5, 2, 0, 0, 1, 4, 0, 6, 5, 3)), (0.738647997379303, 75.36453247070312, (0, 1, 6, 6, 6, 0, 4, 6, 6, 4, 6, 6, 1, 5, 5, 1, 3, 1, 5, 2, 2, 3)), (0.7416280031204223, 75.76422119140625, (0, 0, 6, 6, 6, 2, 0, 2, 3, 4, 6, 1, 6, 0, 1, 1, 1, 5, 4, 2, 3, 3)), (0.7434999942779541, 76.10467529296875, (0, 0, 6, 0, 6, 2, 3, 6, 0, 4, 3, 4, 5, 3, 0, 6, 0, 1, 0, 2, 2, 3)), (0.7470040082931518, 78.22792053222656, (0, 0, 0, 6, 6, 1, 0, 6, 0, 5, 0, 2, 1, 4, 0, 2, 2, 5, 5, 0, 2, 3)), (0.7460600018501282, 80.42455291748047, (0, 0, 6, 0, 6, 2, 0, 0, 2, 1, 0, 0, 6, 5, 0, 1, 3, 5, 5, 2, 2, 3)), (0.7469719886779785, 82.44024658203125, (0, 0, 6, 6, 0, 2, 0, 0, 4, 5, 4, 6, 2, 4, 3, 3, 6, 5, 1, 2, 2, 3)), (0.7474720120429993, 84.94596862792969, (0, 0, 0, 0, 6, 1, 3, 1, 6, 4, 2, 1, 0, 2, 6, 0, 5, 4, 2, 2, 5, 3))]

\end{document}